\pdfoutput=1

\documentclass[11pt]{article}

\usepackage[]{ACL2023}

\usepackage{times}
\usepackage{latexsym}

\usepackage[T1]{fontenc}

\usepackage[utf8]{inputenc}

\usepackage{microtype}

\usepackage[subtle]{savetrees}
\usepackage{inconsolata}
\usepackage{amsmath}
\usepackage{hyperref}
\usepackage{graphicx}
\usepackage{booktabs}
\usepackage{tabularx}
\usepackage{rotating}
\usepackage{pdflscape}
\usepackage{longtable}
\usepackage{multirow}
\usepackage{amssymb}

\usepackage{listings}
\title{Evaluating LLM-Based Grant Proposal Review via Structured Perturbations}

\author{
       William Thorne$^1$,~~~ Joseph James$^1$, ~~~Yang Wang$^2$ \\
\textbf{Chenghua Lin$^{2}$, ~~~ Diana Maynard$^{1}$} \\
     $^1$University of Sheffield, UK~~~~~
    $^2$University of Manchester, UK \\
    \texttt{\{wthorne1,jhfjames1,d.maynard\}@sheffield.ac.uk} \\
    \texttt{\{yang.wang-2,chenghua.lin\}@manchester.ac.uk}}

\begin{document}
\maketitle
\begin{abstract}

As AI-assisted grant proposals outpace manual review capacity in a kind of ``Malthusian trap'' for the research ecosystem, this paper investigates the capabilities and limitations of LLM-based grant reviewing for high-stakes evaluation. Using six EPSRC proposals, we develop a perturbation-based framework probing LLM sensitivity across six quality axes: funding, timeline, competency, alignment, clarity, and impact. We compare three review architectures: single-pass review, section-by-section analysis, and a 'Council of Personas' ensemble emulating expert panels. The section-level approach significantly outperforms alternatives in both detection rate and scoring reliability, while the computationally expensive council method performs no better than baseline. Detection varies substantially by perturbation type, with alignment issues readily identified but clarity flaws largely missed by all systems. Human evaluation shows LLM feedback is largely valid but skewed toward compliance checking over holistic assessment. We conclude that current LLMs may provide supplementary value within EPSRC review but exhibit high variability and misaligned review priorities. We release our code and any non-protected data\footnote{Our code can be found here: \url{https://github.com/wrmthorne/grant-perturbation-analysis}}.
\end{abstract}

\section{Introduction}
\label{section:introduction}

Peer review is the foundational mechanism for ensuring scientific rigour and directing funding towards impactful, feasible research. However, the global research ecosystem is currently caught in a \textit{``Malthusian trap''} \citep{Naddaf2025}: while R\&D funding has seen incremental increases, the volume of applications has grown exponentially. In the UK, the number of competitive grant applications assessed by UK Research and Innovation (UKRI) has nearly doubled since 2017, while the overall award rate has plummeted from 36\% to 19\% \citep{ResearchProfessional2025}. This surge has placed the research ecosystem under unprecedented strain, resulting in systemic reviewer fatigue and a rising administrative burden \citep{tickell2022independent}. Consequently, the peer-review process has seen significantly extended decision cycles, with the end-to-end timeline for major funding schemes now frequently exceeding 18 months \citep{ResearchProfessional2025}. This pressure is exacerbated by a growing \textit{dual-standard} in generative AI (GenAI) policy \citep{Reidpath2024}. While  policy is increasingly permissive for applicants, allowing GenAI for brainstorming, structuring, and language editing, it remains strictly prohibited for reviewers. Allowing applicants but not reviewers to use LLMs creates an asymmetry that risks either lower review quality or longer funding timelines.

Grant evaluation also presents challenges that distinguish it from the more widely studied conference setting. Unlike paper reviewing, which is retrospective (evaluating completed work), grant reviewing is prospective and administrative, requiring high-stakes assessments of value for money, multi-year project feasibility, and national impact. It further demands: (1)~\textbf{Contextual Breadth}, assessing diverse documents from financial spreadsheets to impact statements; (2)~\textbf{Applicant Visibility}, where non-anonymous metadata increases the risk of prestige or institutional bias; and (3)~\textbf{Decision Stakes}, where errors carry consequences associated with significant capital and multi-year commitments. While recent research has explored LLM capabilities in academic peer review for conferences~\citep{liang2024can, AAAI2025}, evaluating their readiness for grants demands a fundamentally different methodology. Complete research proposals are tightly guarded assets whose contents comprise novel and highly valuable intellectual property, both financially and in terms of participant careers and reputations. The scarcity of data and the high ethical barriers to access are central to why grant proposals and their reviewing are so understudied, despite their importance.

Recent work using LLMs in the grant domain has been primarily applicant-focused, assisting with proposal drafting, literature discovery and alignment with funding criteria \citep{seckelTenSimpleRules2024, automatingGrantWriting2025}. \citet{okasa2025supervised} developed a supervised pipeline on the content of grant review reports at the Swiss National Science Foundation; however, this only analyses existing human reviews instead of synthesising or evaluating them. To our knowledge, no prior work has systematically evaluated LLM capabilities for the review of grant proposals: assessing whether models can identify substantive weaknesses, produce reliable scores or generate feedback comparable to expert reviewers. Our work addresses this gap.

Under this premise, we propose \textbf{perturbation-based evaluation} as a principled solution to this issue of data-scarcity. Rather than creating supervision through the labelling of many proposals, we construct controlled fault conditions from a limited pool of genuine grant submissions and measure LLM review systems’ ability to reliably detect known defects. Utilising six genuine Engineering and Physical Sciences Research Council (EPSRC) proposals, we define a perturbation taxonomy based on six key axes of quality (funding, timeline, competency, alignment, clarity, and impact), decomposing into 42 total perturbations at the most granular level. Each of these may be applied to every proposal to produce variants that reflect known, targeted weaknesses relative to the original. We summarise our contributions as follows: \textbf{(1) a perturbation-based evaluation framework} that enables principled, fine-grained assessment of LLM review systems in data-scarce, high-sensitivity domains, demonstrated here by transforming six proposals into 42 controlled fault conditions across six quality axes; \textbf{(2) the development of a Council of Personas architecture} designed to emulate the multi-perspective nature of expert panels; and \textbf{(3) a comparative analysis of model-generated feedback} against the nuanced judgements of experienced UKRI reviewers to identify current gaps in automated reasoning and potential sources of bias.

Specifically, we address three exploratory research questions: (\textbf{RQ1})~How do different architectural configurations influence the detection of systematic proposal perturbations and the reliability of scoring? (\textbf{RQ2})~Which core assessment dimensions are LLM-based systems most sensitive to? (\textbf{RQ3})~How does the qualitative feedback from LLMs align with the judgements of experienced UKRI reviewers?

\section{Related Work}
\label{section:related_work}

\paragraph{LLM-Assisted Peer Review}
Recent studies on LLMs in academic publishing focus on two areas: autonomous review generation and reviewer assistance. While models can identify surface-level reproducibility and formatting issues \citep{neurips2025, iclr2025}, they often struggle with nuanced methodological flaws and novelty detection \citep{du2024llms, zhou2024llm}. Automated systems like \textit{Reviewer3} \citep{Reviewer3} show moderate agreement with humans, but large-scale studies suggest that while LLMs (e.g., GPT-4) overlap with human consensus as much as humans do with each other, the quality of their qualitative feedback remains limited \citep{liang2024can}. However, another study comparing ChatGPT with human reviewers found low levels of agreement, noting in particular that the former could not adjust to the specific requirements or focus of a particular journal \citep{checco2023ai}.

A 2022 report for the four UK HE funding bodies \citep{thelwall2022can} concluded that AI tools are not yet capable of helping to make scoring decisions about journal article scores in future research assessment exercises. However, it recognised the potential, and recommended further exploration through pilot testing. The conclusions of this work predate the current generation of reasoning models; our work provides an updated empirical assessment of whether this potential has been realised.

\paragraph{Grant Reviewing vs. Paper Reviewing}

Although proposals and publications are both assessed according to academic rigour and novelty, the prospective nature of grant applications means that assessment is distinguished by its focus on \textit{justification} and \textit{feasibility} \citep{he2011learning}. Successful proposals must align with specific funding call requirements and demonstrate a clear gap in the current landscape \citep{weidmann2023write}. Previous NLP work in this domain has largely focused on automated classification of reviewer reports \citep{okasa2025supervised} or analysing funding trends, rather than evaluating the generative and critical capabilities of LLMs in the review loop. Existing work consistently shows that while LLMs can summarise and identify specific content in a paper, they often struggle to identify nuanced methodological weaknesses, feasibility and deep insights into novelty \citep{du2024llms,zhou2024llm}. These limitations mirror the core challenges of grant assessment, where evaluating criteria such as team competency and resource justification requires a holistic synthesis of the proposal's vision, extending beyond the surface-level pattern matching typical of current models.

\paragraph{Stance Detection and Argument Mining}
Early work on stance detection focused on social media and debate data, including the SemEval Twitter shared tasks \citep{mohammad-etal-2016-semeval} and cross-target generalisation approaches \citep{augenstein-etal-2016-stance}. Further work extended stance modelling to more structured and knowledge-intensive domains such as news and political discourse, and surveys have synthesised advances in neural and transformer-based stance models \citep{hardalov-etal-2022-survey}. Within scientific communication, related work has examined citation sentiment and scientific disagreement \citep{athar-2011-sentiment, jurgens-etal-2018-measuring}, showing that evaluative language often reflects epistemic positioning rather than simple polarity.

Argument mining provides a complementary perspective by decomposing texts into structured argumentative units. Prior work has introduced frameworks for identifying argument components and their relations \citep{stab-gurevych-2014-annotating, peldszus-stede-2015-joint}. These approaches have been adapted to scientific writing and reviewing, including rhetorical role annotation \citep{liakata-2010-zones} and proposition-level decomposition of peer reviews \citep{kovatchev-etal-2020-decomposing}.  In natural language inference and scientific claim verification \citep{bowman-etal-2015-large, wadden-etal-2020-fact} model agreement and contradiction between pairs of claims, providing a formal basis for claim-level comparison.

\paragraph{Evaluation via Perturbation}
Perturbation-based evaluations have emerged as a standard methodology for probing LLM robustness and the reliability of the ``LLM-as-a-judge'' paradigm \citep{Chaudharyetal2024, hong2025beyond}. By systematically altering input text while maintaining core semantic properties, researchers can identify specific model biases and failure modes that remain hidden during standard benchmarking. While foundational frameworks like TextAttack \citep{morris-etal-2020-textattack} focus primarily on lexical and syntactic transformations, such as word substitutions or character-level noise, recent benchmarks have expanded this to evaluate model consistency across diverse document formats and enterprise-scale contexts \citep{Bogavellietal2026}.

In this work, we extend the scope of perturbation from linguistic variations to domain-specific structural inconsistencies. Unlike general-purpose benchmarks, we target the logical ``fatal flaws'' unique to the grant domain, such as budget-timeline misalignments or competency gaps, to test whether LLMs can move beyond pattern matching toward the rigorous, high-stakes reasoning required for professional research assessment.

\section{Methodology}
\label{section:methodology}

Research grant proposals are highly sensitive assets, containing both proprietary intellectual property and confidential data of the participants. Due to these privacy constraints and the high ethical burden associated with their processing, the systematic study of grant reviewing remains significantly under-researched. To address these challenges, we evaluate the capability of offline, locally-served LLMs to review proposals submitted to EPSRC. Our methodology maximises the utility of a limited dataset through a perturbation-based approach: we begin with a set of contemporary, human-authored proposals, submitted to EPSRC, and systematically degrade their quality. We deconstruct the notion of proposal quality along six axes: \textit{funding}, \textit{timeline}, \textit{competency}, \textit{alignment}, \textit{clarity} and \textit{quality}, and \textit{impact}, which we derive from the UKRI assessment process itself. Through this approach, we demonstrate that this (albeit limited) dataset can nevertheless be augmented into a robust benchmark to test the sensitivity, consistency and capabilities of current LLMs for EPSRC proposal reviewing.

\subsection{Review Frameworks}
\label{section:review_frameworks}

\subsubsection{Zero-shot Baseline}
\label{section:baseline-reviewer}

The baseline system, \textbf{GPT-OSS-20B (\textit{high})} \citep{openai2025gptoss120bgptoss20bmodel}, is provided with a zero-shot task description, the official UKRI review guidelines, the specific funding opportunity, and the complete proposal narrative in a single context. The model is tasked with producing an overall score (1-6), following the official EPSRC reviewer scoring scale, and a set of comments justifying this score and providing feedback. 

\subsubsection{Section-Level Review Framework}
\label{section:section-level-reviewer}

Input prompts to the baseline approach often exceed $30,000$ tokens. While modern LLMs demonstrate high accuracy in simple information retrieval at these lengths, their ability to perform complex reasoning and synthesis decays significantly as context scales \citep{liu-etal-2024-lost, hsieh2024ruler}. This performance gap, where models struggle to apply a fact to a critical evaluation, is particularly acute in document-level assessment \citep{wu2024long}. To mitigate this, we implement a section-level review process, reducing the cognitive load per inference pass encouraging more specific feedback \citep{wu-etal-2024-less}.

We created four logical groups from the sections of the proposal documents as many sections provide little value to review in isolation (e.g. references) but provide valuable context when accompanying others. The groups are: \textbf{Vision-Approach} (vision and approach, references); \textbf{Team Capability} (summary, applicant and team capability to deliver, core team, project partners, facilities, references); \textbf{Funding Resources} (summary, resources and costs, core team, facilities, references); and \textbf{Ethics} (summary, ethics and responsible research and innovation, research involving human participants). We discard purely administrative sections (EPSRC thematic area alignment, letters of support) and sections which were not applicable to any of our reviews (e.g. animal testing, sensitive information).

\subsubsection{Council of Personas}
\label{section:council-reviewer}

The baseline and section-level approaches risk propagating single-perspective biases and linear errors into the final feedback. To address this, we employ a \textbf{Council of Personas}\footnote{We use Andrej Karpathy's repo as reference \url{https://github.com/karpathy/llm-council}.}, implementing majority-voting for qualitative reasoning via a three-stage process: (1) independent persona-based reviews; (2) blind meta-review and ranking of peer councillor outputs; and (3) final synthesis by a council chair who down-weights anomalous comments based on these rankings.

We use five different personas to encourage feedback diversity: \textbf{Cost Analyst, Ethics Assessor, Tech Evangelist, Methodological Sceptic}, and \textbf{Impact Champion}. Each persona introduces a deliberate bias: for instance, the Methodological Sceptic prioritises soundness and validity, while the Impact Champion focuses on scalability and industry engagement. Full persona descriptions and prompting templates are provided in \textbf{Appendix A.1}. This collective approach ensures that while individual personas may be over-sensitive to specific ``fatal flaws,'' the final output remains holistic and aligned with standard UKRI guidance.

\subsection{Data}
\label{sec:data}

Our dataset comprises six full EPSRC funding applications from the School of Computer Science, obtained through collaboration with our institutional research hub. Of these, two proposals were successfully funded, one was unfunded, and the remaining three are awaiting a decision; one funded and one unfunded proposal are accompanied by full expert-review comments and scores. The original submission dates range from May 2023 to August 2025; working with recent proposals mitigates data leakage risks, as it is unlikely this specific content appeared in the pre-training data of current models. Furthermore, as unpublished grant proposals, these documents are not publicly accessible and could not have been included in the training corpora of the models under evaluation.

\paragraph{Assessment Context}
Standard UKRI assessment involves evaluation by 3-4 expert reviewers who score applications on a scale of 1-6 across four primary pillars: \textit{Research Excellence}, \textit{National Importance}, \textit{Applicant Track Record}, and \textit{Resources and Management}. Consequently, our analysis focuses on the \textit{Vision and Approach}, \textit{Team Capability to Deliver}, \textit{Justification of Resources}, and \textit{Ethics} sections, as these contain the primary claims where ``fatal flaws'' in feasibility typically appear.

\paragraph{Preprocessing}
For each proposal, we obtain relevant contextual data including the target opportunity and, where available, public project data from Gateway to Research (GtR). All textual content is converted to markdown to preserve structural features such as bolding and header nesting. We process PDFs using Docling \citep{docling} followed by manual cleaning to reintroduce hyperlink markup.

To handle the critical timeline data often trapped in images, we explored several serialization strategies. While Mermaid and PlantUML render well for humans, their syntax bears little resemblance to the visual timeline. We therefore adopted a markdown table syntax with cells shaded using ``\#\#\#\#'', which proved effective for LLM consumption and facilitated the systematic perturbations described in Section~\ref{section:perturbations}.

\paragraph{Perturbation Strategy}
\label{section:perturbations}

To evaluate model sensitivity across the core dimensions of the UKRI assessment process, we systematically degrade the proposals along six primary axes. Table \ref{tab:perturbation-taxonomy} summarises the perturbation strategies employed and the specific evaluation criteria they target.

Our perturbation strategy was informed by an initial round of human evaluation. Four existing members of the EPSRC review college, each with many years of EPSRC grant reviewing experience, were given either the vision or approach section. They were tasked with scoring the section between one and six, following the standard UKRI rubric and justify their score using positive and negatively highlighted excerpts from the text and a separate, discursive set of positive and negative feedback.

Evaluators noted that referencing was generally a strength, yet not always sufficiently developed to make limitations, motivations and novelty explicit (e.g. Implications of novelty but never fully articulated against prior work). At the same time, information could become overly dense, especially in technical sections, making it difficult to flow through the proposal without backtracking (several mentions of lack of explanation of specific acronyms and other background information). Though evaluators were not domain experts on each proposal, all were within computer science. Clear organisation, explicit sectioning, named references and the use of examples were identified as features that improved clarity and accessibility. These findings led us to the clarity axes, which capture weaknesses in framing, evidential sufficiency and academic rigour.

The evaluation also revealed disconnects between components within sections. Claims of timeliness were not always supported by factual evidence, and the relationship between timeliness, impact and motivation were sometimes fragmented leading to the logical flow being disrupted. Speculative reasoning was often required for evaluators to understand motivation and feasibility, such as lack of preliminary experiments or studies. These factors informed the timeline and impact axes to look at feasibility, justification and contributions of a proposal.

Evaluators also highlighted the importance of alignment between resources, expertise and positioning of the proposed work. Proposals could appear weaker when different components did not clearly reinforce one another, or when justification in one section was not adequately supported elsewhere. Even where individual sections were strong, a lack of clear connection across components reduced overall credibility. These broader concerns informed the Funding, Competency and Alignment axes.

\begin{table*}[t]
\centering
\scriptsize 
\begin{tabularx}{\textwidth}{l >{\hsize=1.35\hsize}X >{\hsize=0.65\hsize}X}
\toprule
\textbf{Axis} & \textbf{Perturbation Strategy} & \textbf{Targeted Criterion} \\
\midrule
\textbf{Funding} & Inflating/lowering budgets; removing cost justifications; misaligning resource allocation with project priorities. & Value for money; compliance with UKRI financial policy. \\
\addlinespace
\textbf{Timeline} & Extending periods beyond call limits; unrealistic task compression; misaligning milestones with logical work progression. & Project feasibility; operational justification. \\
\addlinespace
\textbf{Competency} & Removing/replacing key personnel; weakening evidence of technical skills in the \textit{team capability} section. & Demonstration of requisite skills and leadership. \\
\addlinespace
\textbf{Alignment} & Modifying opportunity aims; switching ``What we're looking for'' sections; introducing cross-disciplinary mandates. & Strategic fit to call; adherence to funding body values. \\
\addlinespace
\textbf{Clarity} & Removing acronym expansions; introducing vagueness in methods; removing novelty markers; deleting factual background. & Accessibility; technical comprehensibility; academic rigour; specificity. \\
\addlinespace
\textbf{Impact} & Replacing key stakeholders with irrelevant parties; modifying outcome scope; removing long-term/short-term outcomes. & Contribution to field; stakeholder engagement. \\
\bottomrule
\end{tabularx}
\caption{Summary of perturbation axes, strategies, and their mapping to EPSRC assessment criteria. Examples and further details can be found in \autoref{appendix:perturbation-examples} and \autoref{app:taxonomy}.}
\label{tab:perturbation-taxonomy}
\end{table*}

\section{Experimental Setup}
\label{section:experimental_setup}

\subsection{Evaluation Tasks}
\label{section:evaluation-tasks}

We evaluate our systems using two primary tasks. To simulate secure deployment and satisfy data privacy requirements, all experiments were conducted on isolated single-GPU devices without external access. We used the \textit{InspectAI}\footnote{\url{https://inspect.aisi.org.uk/}} python library for repeatability and extensibility, adopting its standard terminology throughout our evaluation.

\subsubsection{Perturbation Identification}
\label{section:perturbation-identification}

We assess the sensitivity of LLM review systems to different aspects of quality by introducing repeatable and independent perturbations into our human-authored grant proposals. Each perturbation was designed to have a solely negative impact on one of the six axes of quality. Sensitivity is measured by observing the proportion of review comments that negatively address the perturbation (e.g. no explanation of abbreviations) or an obvious direct consequence of it (e.g. the section is unclear as none of the methods are explained). Responses are scored as correct only if the perturbation or consequence was both mentioned and with negative sentiment. In cases where the comment partially addresses the perturbation, a consequence could have arisen by multiple means or in any other cases of ambiguity, partial credit is assigned. Contradictory cases of clear identification but positive sentiment  are deemed incorrect.

To scale the analysis to all 42 perturbations across all six proposals and in each review setting, we employed a panel of three judge models: Qwen3.5-35B-A3B \citep{qwen3.5}, NVIDIA-Nemotron-3-Nano-30B-A3B \citep{nvidia_nemotron_nano_v3_2025}, and GLM-4.7-Flash \citep{5team2025glm45agenticreasoningcoding}. Each judge independently evaluates whether the review identifies the perturbation, and we take the majority verdict. Each judge is given a human-authored description of the perturbation, along with the file diffs between the original and perturbed proposals. The judge first engages in a reasoning phase before responding with one of Correct (C), Partial (P) or Incorrect (I). Providing this list of differences as context mitigates any misinterpretation from our descriptions, while not polluting the context. The prompt template can be found in~\ref{section:judge-prompt}.

To validate the judge panel, we assess inter-annotator agreement between the three models. The panel achieves a Krippendorff's $\alpha$ of 0.74, indicating substantial agreement. We also perform a comparison against a human-annotated subset of 50 samples in which we find Qwen3.5 and GLM-4.7 to exhibit perfect agreement with the human annotator ($kappa = 1.0$), while Nemotron shows a fair alignment of $\kappa = 0.57$. The majority voting mechanism effectively compensates for Nemotron's higher rate of disagreement, resulting in reliable verdicts.

This task allows us to explore how architectural choices affect detection performance, which assessment pillars LLMs can and cannot identify, and whether their feedback aligns with expert judgment.

\subsubsection{Expert–Model Feedback Alignment}
\label{section:setup-expert-model-feedback-alignment}

Review comments provide applicants with critical feedback for resubmission and offer transparency into the decision-making process. We reframe these reviews as sets of atomic claims, each with a valence (positive, neutral, or negative), comparing model-generated claims ($M$) to expert reviews ($E$). Their intersection ($M \cap E$) represents consensus, while claims unique to experts ($E \setminus M$) identify areas where the model lacks technical depth or alignment with reviewer priorities. Finally, valid claims exclusive to the model ($M \setminus E$) represent the potential additive value of LLMs, highlighting insights that human reviewers may have overlooked.

We perform a human annotation of LLM claims that do not appear in expert claims. The two proposals we received with review comments each have four original expert reviews. Following a manual assessment, we observe no direct contradictions in the claims made between human reviewers for each proposal. Contradictions are defined as factual discrepancies or opposing valence between claims, such that the claims cannot both be true at the same time \citep{kovatchev-etal-2020-decomposing}.

We generate reviews for the two proposals using each of our LLM review systems and extract individual claims from them. To extract atomic claims, we utilise GPT-OSS-120B (high) to break down review sentences into their atomic form. We define a review claim as a claim referring to a specific topic within the grant that expresses a positive, neutral or negative valence, thereby capturing constructive feedback, identified strengths and weaknesses \citep{kovatchev-etal-2020-decomposing}. In many cases, a single claim sentence can contain multiple evaluative aspects, for example assessing feasibility and novelty simultaneously. Decomposing such sentences into atomic units isolates each evaluative component as a separate claim.

We therefore created a taxonomy to split claims into specific categories that overlap with proposal sections (see \autoref{app:taxonomy}). Once a claim is labelled with a taxonomy, semantically similar claims are grouped using embedding-based clustering with Qwen3-Embedding-8B \citep{qwen3embedding}. This allows related claims to be grouped before assigning an aspect. For each resulting cluster, the model generates a concise three-word aspect that captures the cluster's topic, reducing variability in aspect labels across similar claims.

Decomposed claims sharing the same source sentence, valence, and aspect are re-merged into their original composite form; otherwise they remain decomposed. After processing, we perform bi-directional relevance matching using E2Rank \citep{liu2025e2rank}, an embedding-to-rank model combining retrieval and listwise re-ranking. For each claim in the LLM set, we retrieve the most similar claims from the human set above a similarity threshold of 0.5 as potential matches, and vice versa.

Each claim, together with its associated potential matches, is then labelled as EXACT, DIFFERENT or CONTRADICTION, following a similar framework to \citet{fritsch2025callm}. EXACT indicates that the LLM and human claims address the same topic and express the same valence, while CONTRADICTION indicates alignment in topic but opposing valence. DIFFERENT captures partial overlap in topic without clear agreement in evaluative valence. A further human quality control step was conducted to remove incomplete, overly generic claims introduced during decomposition. The final dataset consists of 500 claims, and the full breakdown is shown in \autoref{tab:human_eval_breakdown}.

\begin{table*}[!t]
\centering
\footnotesize
\begin{tabular}{l ccc ccc ccc}
\toprule
& \multicolumn{3}{c}{\textbf{Baseline}} & \multicolumn{3}{c}{\textbf{Section-Level}} & \multicolumn{3}{c}{\textbf{Council}} \\
\cmidrule(lr){2-4} \cmidrule(lr){5-7} \cmidrule(lr){8-10}
\textbf{Perturbation Axis} & C & P & I & C & P & I & C & P & I \\
\midrule
Funding (n=1688)     & 159 & 20 & 440 & 220 & 40 & 358 & 115 & 51 & 285 \\
Competency (n=889)   & 28 & 12 & 353 & 32 & 10 & 240 & 15 & 7 & 192 \\
Alignment (n=953)    & 122 & 2 & 222 & 183 & 1 & 163 & 84 & 0 & 176 \\
Clarity (n=1859)     & 6 & 0 & 668 & 99 & 5 & 571 & 12 & 4 & 494 \\
Impact (n=639)       & 10 & 5 & 221 & 71 & 5 & 157 & 7 & 0 & 163 \\
Timeline (n=1088)    & 85 & 0 & 314 & 106 & 3 & 289 & 40 & 2 & 249 \\
\midrule
\textbf{Overall} (n=7116) & 410 & 39 & 2218 & 711 & 64 & 1778 & 273 & 64 & 1559 \\
\bottomrule
\end{tabular}
\caption{Perturbation identification rates across review systems. C = Correct (perturbation identified with negative sentiment), P = Partial (ambiguous or indirect identification), I = Incorrect (missed or positive sentiment). Verdicts are determined by majority vote across three independent judge models.}
\label{tab:perturbation-results}
\end{table*}

For human evaluation claims are given per section of the proposal due to ethical limitations of sharing personal data and intellectual property outside of the research team. Each annotator reads a section, opportunity and the review guidelines. Each task is a separate claim that they must state 1) the validity of the claim, 2) their agreement with (strong disagree, disagree, neutral, agree, strong agree) and 3) rate the significance of the comment with respect to how much impact that claim would have on the review score, assuming it was correct.

Within human reviewing, an absence of comment can often be viewed as a satisfaction of the requirements. Trivial additional positive comments do not indicate significant value. Similarly human reviewers might list only the most significant issues necessary to kill a proposal. Minor negative comments missing from human reviews indicate potential for value in delivering constructive criticism and actionable feedback to applicants. Major negatives absent from human reviews indicate actual opportunity for value that could be provided by an LLM.

\subsection{Metrics}
\label{section:metrics}

We employ the following metrics to evaluate review system performance:

\paragraph{Perturbation Detection Score.}~
A numerical mapping from the judge's verdict: Correct = 1.0, Partial = 0.5, Incorrect = 0.0. This enables continuous analysis of detection performance.
\vspace{-0.5em}

\paragraph{Score Degradation.}~
The signed difference $\Delta S = S_{\text{original}} - S_{\text{perturbed}}$ between scores assigned to original and perturbed proposals. Positive values indicate appropriate score reduction; negative values flag anomalous cases where perturbation \textit{increased} the score.
\vspace{-0.5em}

\paragraph{Intra-Class Correlation (ICC).}~
We use ICC(2,1) to assess scoring reliability, treating both proposals and evaluation runs as random effects. This two-way random effects model quantifies the proportion of variance attributable to true differences between proposals:
\vspace{-0.5em}

\begin{equation}
\text{ICC}(2,1) = \frac{\sigma^2_p}{\sigma^2_p + \sigma^2_r + \sigma^2_e}
\end{equation}
where $\sigma^2_p$ is variance due to proposals, $\sigma^2_r$ is variance due to raters/runs, and $\sigma^2_e$ is residual error. Higher ICC indicates greater reliability, with values above 0.5 considered moderate \cite{kooGuidelineSelectingReporting2016a}.

\section{Results and Discussion}
\label{section:results}

\subsection{Perturbation Identification}
\label{section:results-perturbation}

We evaluated 42 unique perturbations across 6 proposals using 3 review systems (7,347 perturbed observations total). The overall detection rate was 21.2\%; nearly four in five perturbations go undetected.

\paragraph{System Comparison.}
Table~\ref{tab:perturbation-results} presents detection rates, aggregated by perturbation type. The section-level review system achieved the highest detection scores across nearly all categories ($\mu = 0.29$), followed by the baseline ($\mu = 0.17$) and council approach ($\mu = 0.17$). We confirm that the performance difference between systems is significant via a Kruskal-Wallis\footnote{We note that our scores are ordinal and subsequently mapped to floats} test ($H = 27.62, p < 0.0001$)~\cite{kruskal_wallace}.

Through a pairwise comparison of the systems, we find the baseline and council methods produce scores that are statistically indistinguishable ($p=0.83$); this is an especially poor result for the council setup given the token cost. Conversely, we observe that the section-level system consistently prescribes lower scores than the other two with a mean difference of around 1.2 points ($p < 10^{-46}$). This could be seen as miscalibration or an overly harsh reviewer; however, the high detection rate suggests a more critical and accurate reviewer.

\begin{figure}
    \centering
    \includegraphics[width=\linewidth]{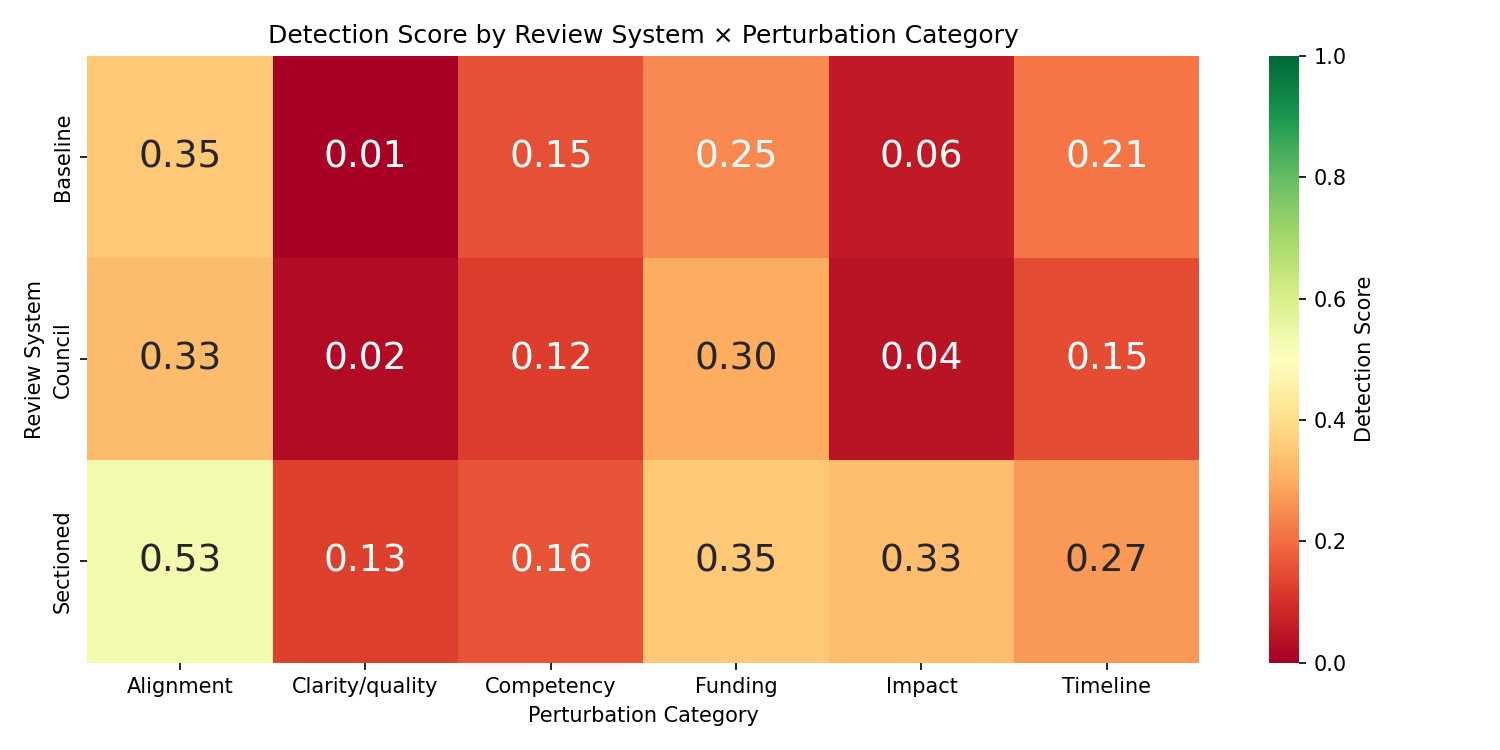}
    \caption{Detection scores across review systems and perturbation categories. Darker cells indicate higher detection rates. The section-level system shows strongest performance on alignment and impact perturbations, while all systems fail on clarity-based changes.}
    \label{fig:detection-heatmap}
\end{figure}

\paragraph{Perturbation Sensitivity.}
Detection rates varied greatly across different perturbation categories. Perturbations to the alignment were most detectable ($\mu = 0.41$), particularly the \textit{cross-cutting theme injections} ($\mu = 0.70$); however, We note that the alignment perturbations were performed on the opportunity documents rather than the proposals themselves. We believe that, because many opportunity documents likely appear in models' pre-training data, they have learned the typical structure and conditions of opportunity notices. Deviations from these patterns may appear more salient than comparable changes within proposals which have never been seen. Clarity perturbations, on the other hand, clarity went almost entirely undetected ($\mu = 0.06$); \textit{acronym-related} changes and \textit{connective removal} were never identified. Figure~\ref{fig:detection-heatmap} demonstrates the interaction between review system and perturbation type. We attribute the consistently poor performance on clarity in part to the subtlety of these perturbations given their subtlety; however, we believe the LLM review systems rely on contextual inference to resolve ambiguous terminology or acronyms rather than flagging them as missing definitions or quality concerns. While this should be a strength, we observe an over-reliance on this ability, with LLM review systems failing to question random abbreviations in the text.

\paragraph{Reliability.}
We decompose the total variance in scores to assess the consistency of each review system (Table~\ref{tab:variance:decomposition}). The section-level approach achieves the highest intra-class correlation (ICC = 0.50), indicating that approximately half of the observed variance reflects true differences between proposal versions rather than noise. In contrast, both the baseline (ICC = 0.14) and council (ICC = 0.11) systems exhibit substantially higher within-sample variance, meaning repeated evaluations of the same proposal yield inconsistent scores. The council's poor reliability is particularly notable given its significantly higher computational cost (see Appendix); the multi-persona architecture does not translate into more stable assessments. These findings suggest that decomposing the review task into focused sections yields more reproducible judgments than either holistic processing or ensemble-based approaches.

\begin{table}
    \footnotesize
    \centering
    \begin{tabular}{c|cccc}
        {} & $\sigma^2_{\text{total}}$ & $\sigma^2_{\text{between}}$ & $\sigma^2_{\text{within}}$ & ICC ($\uparrow$) \\
        \hline
        Baseline & 0.87 & 0.13 & 0.80 & 0.14 \\
        Council & 0.91 & 0.11 & 0.88 & 0.11 \\
        Section-Level & 0.88 & 0.49 & 0.49 & \textbf{0.50}
        \\
    \end{tabular}
    \caption{Variance decomposition across review systems. ICC (intra-class correlation) measures the proportion of variance attributable to true differences between proposals versus noise from repeated evaluation. }
    \label{tab:variance:decomposition}
\end{table}

\subsection{Expert–Model Feedback Alignment}
\label{subsection:results-expert-model-feedback-alignment}

\begin{table*}[t!]
\centering
\footnotesize
\renewcommand{\arraystretch}{1.2}
\setlength{\tabcolsep}{6pt}
\begin{tabular}{lcccccccc}
\toprule
& \multicolumn{2}{c}{\textbf{Human}}
& \multicolumn{2}{c}{\textbf{Baseline}}
& \multicolumn{2}{c}{\textbf{Council}}
& \multicolumn{2}{c}{\textbf{Section-Level}} \\
\textbf{Category}
& \textbf{Ret} & \textbf{Ctr}
& \textbf{Ret} & \textbf{Ctr}
& \textbf{Ret} & \textbf{Ctr}
& \textbf{Ret} & \textbf{Ctr} \\
\midrule
Alignment  & 79.6\% & 2.8\% & 68.2\% & 0.0\% & 71.8\% & 1.9\% & 80.0\% & 0.0\% \\
Competency & 67.7\% & 2.1\% & 76.8\% & 1.8\% & 66.7\% & 1.6\% & 52.2\% & 0.0\% \\
Ethics     & 97.7\% & 0.0\% & 95.0\% & 0.0\% & 94.0\% & 1.2\% & 90.0\% & 0.0\% \\
Funding    & 81.3\% & 8.0\% & 84.0\% & 0.0\% & 89.5\% & 4.8\% & 100.0\% & 0.0\% \\
Impact     & 77.8\% & 1.2\% & 85.3\% & 0.0\% & 73.8\% & 1.0\% & 66.7\% & 0.0\% \\
Clarity    & 93.5\% & 1.6\% & 98.5\% & 0.0\% & 83.9\% & 1.9\% & 86.2\% & 6.9\% \\
Timeline   & 96.7\% & 0.0\% & 100.0\%& 0.0\% & 84.6\% & 2.6\% & 85.7\% & 14.3\% \\
\midrule
\textbf{TOTAL}
& \textbf{82.2\%} & \textbf{2.2\%}
& \textbf{86.5\%} & \textbf{0.4\%}
& \textbf{78.2\%} & \textbf{2.1\%}
& \textbf{77.4\%} & \textbf{2.8\%} \\
\bottomrule
\end{tabular}
\caption{Retained (Ret) and Contradiction (Ctr) Rates of claims.
Retained rate is computed from the \textbf{exclusive set}, where consensus claims are removed, isolating claims unique to each reviewer. Higher retained percentages indicate greater reviewer-specific contribution (unique claims), whereas lower values suggest substantial overlap }
\label{tab:exclusive_contradiction_divergence}
\end{table*}

\begin{figure}
    \centering
    \includegraphics[width=0.65\linewidth]{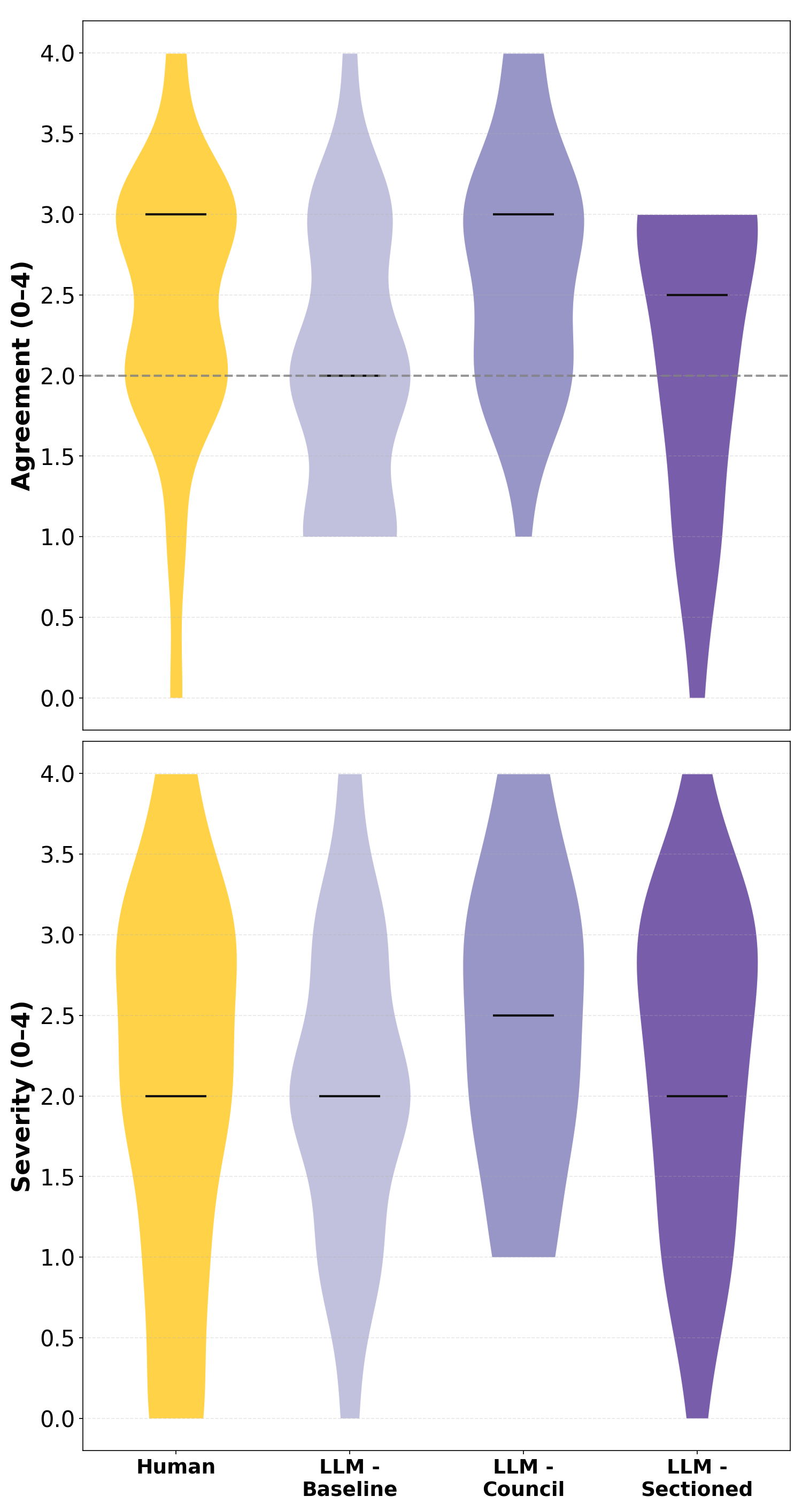}
    \caption{\autoref{tab:claim-influence} shows the scale for severity. The dashed line on agreement indicates the neutral agreement. }
    \label{fig:human_boxplot}
\end{figure}

\paragraph{Human annotation.}~
For the validity label, annotators were asked to label each claim, originating from human reviewers or LLM systems, as either valid or invalid. As this label functions as a quality check, the majority of claims are expected to be valid, resulting in a naturally high percent agreement of 89.5\%. In this setting, Fleiss' $\kappa$ is an unsuitable measure as it penalises agreement that arises from a genuine class imbalance rather than annotator bias \citep{byrt1993bias}.  For Agreement and impact, we see a Fleiss' $\kappa$ of 0.78 and 0.68, respectively, indicating substantial agreement.

As shown in \autoref{fig:human_boxplot}, all systems score above the neutral midpoint for agreement, though human and Council claims show tighter distributions at higher values. For severity, LLM systems tend to prioritise high-severity claims, while human reviewers contribute a broader range, including more low- or "None"-severity observations that provide contextual framing. The Baseline diverges from this pattern, producing a more balanced distribution centred on moderate severity.

\begin{table}[t]
\centering
\footnotesize
\renewcommand{\arraystretch}{0.9}
\setlength{\tabcolsep}{4pt}
\begin{tabular}{llcc}
\toprule
 & \textbf{Review} & \textbf{Spearman $\rho$} & \textbf{Significance} \\
\midrule
\multirow{4}{*}{Agreement}
& Human     & 0.258 & 0.014 \\
& Baseline  & 0.431 & $<0.001$ \\
& Council   & 0.732 & $<0.001$ \\
& Section-Level & 0.613 & 0.006 \\
\midrule
\multirow{4}{*}{Severity}
& Human     & 0.250 & 0.018 \\
& Baseline  & 0.306 & 0.008 \\
& Council   & 0.236 & 0.270 \\
& Section-Level & 0.318 & 0.141 \\
\bottomrule
\end{tabular}
\caption{Spearman correlations between valence and outcome measures by method.}
\label{tab:valence_spearman_by_outcome}
\end{table}

\begin{table}[t]
\centering
\footnotesize
\renewcommand{\arraystretch}{0.9}
\setlength{\tabcolsep}{6pt}
\begin{tabular}{lccc}
\toprule
 & \textbf{Negative} & \textbf{Neutral} & \textbf{Positive} \\
\midrule
Human         & 17.4\% & 33.1\% & 49.5\% \\
Baseline  & 41.7\% & 20.8\% & 37.5\% \\
Council   & 22.2\% & 25.9\% & 52.0\% \\
Section-Level   & 48.1\% & 12.3\% & 39.6\% \\
\bottomrule
\end{tabular}
\caption{Valence distribution (percentage of total claims per source).}
\label{tab:valence_distribution}
\end{table}

\paragraph{Claim analysis.}~
In \autoref{tab:exclusive_contradiction_divergence}, exclusive claim rates are consistently high across all reviewer types, indicating substantial divergence in the issues each raises. Human reviewers show strong exclusivity in Ethics, Clarity, and Timeline, while the baseline exhibits even higher overall exclusivity, particularly in Clarity, Timeline, and Impact, suggesting it frequently introduces points beyond those raised by humans. The council shows slightly lower exclusivity in Competency and Impact, indicating greater overlap, while the section-level model displays more variability across dimensions. Critically, claims rarely contradict each other, averaging around 2\% of total claims, indicating that unique claims are generally additive rather than conflicting. However, uniqueness alone does not guarantee quality, as a unique claim may still be low severity or disagreeable.
Examining individual claims through \autoref{tab:human_eval_breakdown}, the most pronounced variance appears in Ethics, where LLM systems raise specific concerns such as data governance, GDPR compliance, and environmental sustainability, whereas human reviewers typically offer broader assessments affirming that RRI considerations are adequate. This suggests LLMs surface granular compliance criteria that human reviewers either overlook or implicitly accept as satisfied earlier in the funding cycle.

\paragraph{Valence Analysis.}
We examined correlations between valence and human evaluations for both human and LLM claims. Valence was positively correlated with agreement across all systems (Table~\ref{tab:valence_spearman_by_outcome}), though the correlation was small for human claims and substantially stronger for LLM systems, particularly the council and section-level variants. These patterns align with the valence distributions in Table~\ref{tab:valence_distribution}: human and council claims were predominantly positive, whereas baseline and section-level systems generated more negative claims. Valence-severity correlations were weaker and not statistically significant for the council and section-level systems.

Score distributions (Figure~\ref{fig:human_boxplot}) show human and council systems clustering at the upper end of the agreement scale, while baseline and section-level outputs remain more neutral. Human claims exhibit a broader range of severity, reflecting a tendency toward general, affirmatory acknowledgments of proposal content alongside critique rather than focusing exclusively on weaknesses, which accounts for the high variance in severity scores.
These findings suggest that while all systems show a positive correlation between valence and agreement, the stronger effect in LLM systems indicates that models are more proficient at generating positive claims than identifying genuine weaknesses, with LLM-generated criticisms consistently less aligned with human judgment.

\subsection{Overall Discussion}
Our findings highlight that LLMs show genuine but uneven promise for grant reviewing, falling short of the rigour required for a fully automated process. Addressing \textbf{RQ1}, the section-level system consistently outperforms both the baseline and council across detection rates and scoring reliability, indicating that decomposing long context into focused sections is more beneficial than increasing token throughput or architectural complexity alone. Addressing \textbf{RQ2}, sensitivity varies considerably between perturbation types; alignment perturbations are identified at a substantially higher rate than others, whereas clarity perturbations, among the most common concerns raised by human reviewers, go almost entirely undetected. This reflects how LLMs are trained to fill in gaps and resolve ambiguity rather than question it, meaning they are less likely to flag unclear terminology. Addressing \textbf{RQ3}, LLM-generated feedback is largely valid and non-contradictory, but skews toward granular compliance concerns rather than the broader, panel-oriented assessments of experienced UKRI reviewers, surfacing issues such as data governance that human reviewers consider below the threshold of panel relevance. This highlights a fundamental difference between grant and paper peer review: where the latter rewards technical depth, grant reviewing demands holistic judgment about whether a proposal merits public investment.

\section{Conclusion}
\label{section:conclusion}

This paper presents an exploratory investigation into the capabilities of LLMs and their potential for use within the EPSRC grant proposal review process. Using a perturbation-based evaluation framework across six axes of grant quality and a human annotation study with members of the EPSRC review college, we show that current LLM systems exhibit highly uneven sensitivity. Alignment perturbations applied to opportunity documents are identified relatively reliably, likely reflecting internalised patterns from pre-training, whereas clarity-based perturbations to proposals are largely missed.

Overall, current LLMs show significant limitations for autonomous grant review but may offer value as assistive tools within the processes, particularly for structured feedback and alignment checking under human oversight.

\section*{Limitations}

We evaluate six proposals from a single institution using one model family (GPT-OSS), restricting generalisability across disciplines, funding bodies, and architectures. The scale is sufficient for exploratory analysis but precludes strong statistical claims. The ecological validity of our perturbations also varies: some reflect plausible errors (e.g., budget inflation) while others serve as stress tests (e.g., acronym substitution), so detection rates should be interpreted as upper bounds on sensitivity. Finally, our human evaluation relies on reviewers from the same institution as the proposal authors, and the per-section annotation design, while ethically necessary, prevents assessment of cross-section coherence.

\section*{Ethics Statement}

This project has been granted ethical approval by the affiliate institution. All proposals were sourced voluntarily from within the affiliated institution and all handling and processing was conducted on institutional infrastructure. Annotations were conducted also within the institution and in person on an offline device. Annotations were conducted on a per-section level, ensuring any annotator never saw all sections from any single proposal to limit the possibility of plagiarism. All annotators were remunerated at a rate of £25/h.

\section*{Acknowledgements}

This work was supported by the Arts and Humanities Research Council [grant number AH/X004201/1]. Joseph James was supported by the UKRI AI Centre for Doctoral Training in Speech and Language Technologies (SLT) and their Applications funded by UK Research and Innovation [grant number EP/S023062/1].

\bibliography{anthology,custom}

@article{byrt1993bias,
    title = {Bias, prevalence and kappa},
    author = {Byrt, Ted and Bishop, Janet and Carlin, John B},
    journal = {Journal of clinical epidemiology},
    volume = {46},
    number = {5},
    pages = {423--429},
    year = {1993},
    publisher = {Elsevier}
}

@article{tickell2022independent,
    title = {Independent review of research bureaucracy},
    author = {Tickell, A},
    journal = {Recuperado de https://www.gov.uk/government/publications/review-of-research-bureaucracy},
    year = {2022}
}

@inproceedings{fritsch2025callm,
    title = {CALLM: A Framework for Systematic Contrastive Analysis of Large Language Models},
    author = {Fritsch, Reinhard and Jatowt, Adam},
    booktitle = {Proceedings of the 34th ACM International Conference on Information and Knowledge Management},
    pages = {6634--6638},
    year = {2025}
}

@inproceedings{wu-etal-2024-less,
    title = "Less is More for Long Document Summary Evaluation by {LLM}s",
    author = "Wu, Yunshu  and
      Iso, Hayate  and
      Pezeshkpour, Pouya  and
      Bhutani, Nikita  and
      Hruschka, Estevam",
    editor = "Graham, Yvette  and
      Purver, Matthew",
    booktitle = "Proceedings of the 18th Conference of the European Chapter of the Association for Computational Linguistics (Volume 2: Short Papers)",
    month = mar,
    year = "2024",
    address = "St. Julian{'}s, Malta",
    publisher = "Association for Computational Linguistics",
    url = "https://aclanthology.org/2024.eacl-short.29/",
    doi = "10.18653/v1/2024.eacl-short.29",
    pages = "330--343"
}

@article{liu-etal-2024-lost,
    title = "Lost in the Middle: How Language Models Use Long Contexts",
    author = "Liu, Nelson F.  and
      Lin, Kevin  and
      Hewitt, John  and
      Paranjape, Ashwin  and
      Bevilacqua, Michele  and
      Petroni, Fabio  and
      Liang, Percy",
    journal = "Transactions of the Association for Computational Linguistics",
    volume = "12",
    year = "2024",
    address = "Cambridge, MA",
    publisher = "MIT Press",
    url = "https://aclanthology.org/2024.tacl-1.9/",
    doi = "10.1162/tacl_a_00638",
    pages = "157--173"
}

@misc{openai2025gptoss120bgptoss20bmodel,
    title = {gpt-oss-120b \& gpt-oss-20b Model Card},
    author = {OpenAI},
    year = {2025},
    eprint = {2508.10925},
    archivePrefix = {arXiv},
    primaryClass = {cs.CL},
    url = {https://arxiv.org/abs/2508.10925},
}

@article{hong2025beyond,
  title={Beyond one-size-fits-all: Inversion learning for highly effective nlg evaluation prompts},
  author={Hong, Hanhua and Xiao, Chenghao and Wang, Yang and Liu, Yiqi and Rong, Wenge and Lin, Chenghua},
  journal={arXiv preprint arXiv:2504.21117},
  year={2025}
}

@misc{liu2025e2rank,
    title = {E2Rank: Your Text Embedding can Also be an Effective and Efficient Listwise Reranker},
    author = {Qi Liu and Yanzhao Zhang and Mingxin Li and Dingkun Long and Pengjun Xie and Jiaxin Mao},
    year = {2025},
    eprint = {2510.22733},
    archivePrefix = {arXiv},
    primaryClass = {cs.CL},
    url = {https://arxiv.org/abs/2510.22733},
}

@article{qwen3embedding,
    title = {Qwen3 Embedding: Advancing Text Embedding and Reranking Through Foundation Models},
    author = {Zhang, Yanzhao and Li, Mingxin and Long, Dingkun and Zhang, Xin and Lin, Huan and Yang, Baosong and Xie, Pengjun and Yang, An and Liu, Dayiheng and Lin, Junyang and Huang, Fei and Zhou, Jingren},
    journal = {arXiv preprint arXiv:2506.05176},
    year = {2025}
}

@inproceedings{he2011learning,
    title = {Learning the funding momentum of research projects},
    author = {He, Dan and Parker, DS},
    booktitle = {Pacific-Asia Conference on Knowledge Discovery and Data Mining},
    pages = {532--543},
    year = {2011},
    organization = {Springer}
}

@article{weidmann2023write,
    title = {How to write a successful grant application: guidance provided by the European Society of Clinical Pharmacy},
    author = {Weidmann, Anita E and Cadogan, Cathal A and Fialov{\'a}, Daniela and Hazen, Ankie and Henman, Martin and Lutters, Monika and Okuyan, Betul and Paudyal, Vibhu and Wirth, Francesca},
    journal = {International journal of clinical pharmacy},
    volume = {45},
    number = {3},
    pages = {781--786},
    year = {2023},
    publisher = {Springer}
}

@article{okasa2025supervised,
    title = {A supervised machine learning approach for assessing grant peer review reports},
    author = {Okasa, Gabriel and de Le{\'o}n, Alberto and Strinzel, Michaela and Jorstad, Anne and Milzow, Katrin and Egger, Matthias and M{\"u}ller, Stefan},
    journal = {Quantitative Science Studies},
    volume = {6},
    pages = {1189--1214},
    year = {2025},
    publisher = {MIT Press 255 Main Street, 9th Floor, Cambridge, Massachusetts 02142, USA~…}
}

@misc{neurips2025,
    title = {Results of the NeurIPS 2024 Experiment on the Usefulness of LLMs as an Author Checklist Assistant for Scientific Papers},
    author = {NeurIPS},
    year = {2025},
    note = {Accessed: 3 December 2025},
    url = {https://blog.neurips.cc/2024/12/10/results-of-the-neurips-2024-experiment-on-the-usefulness-of-llms-as-an-author-checklist-assistant-for-scientific-papers/}
}

@misc{AAAI2025,
    title = {{AAAI Launches AI-Powered Peer Review Assessment System}},
    author = {AAAI},
    year = {2025},
    note = {Accessed: 3 December 2025},
    url = {https://aaai.org/aaai-launches-ai-powered-peer-review-assessment-system/}
}

@misc{Reviewer3,
    title = {Reviewer3},
    author = {Reviewer3},
    year = {2025},
    note = {Accessed: 3 December 2025},
    url = {https://reviewer3.com/}
}

@misc{iclr2025,
    title = {{Assisting ICLR 2025 reviewers with feedback}},
    author = {ICLR},
    year = {2025},
    note = {Accessed: 3 December 2025},
    url = {https://blog.iclr.cc/2024/10/09/iclr2025-assisting-reviewers/}
}

@inproceedings{du2024llms,
    title = "{LLM}s Assist {NLP} Researchers: Critique Paper (Meta-)Reviewing",
    author = "Du, Jiangshu  and
      Wang, Yibo  and
      Zhao, Wenting  and
      Deng, Zhongfen  and
      Liu, Shuaiqi  and
      Lou, Renze  and
      Zou, Henry Peng  and
      Narayanan Venkit, Pranav  and
      Zhang, Nan  and
      Srinath, Mukund  and
      Zhang, Haoran Ranran  and
      Gupta, Vipul  and
      Li, Yinghui  and
      Li, Tao  and
      Wang, Fei  and
      Liu, Qin  and
      Liu, Tianlin  and
      Gao, Pengzhi  and
      Xia, Congying  and
      Xing, Chen  and
      Jiayang, Cheng  and
      Wang, Zhaowei  and
      Su, Ying  and
      Shah, Raj Sanjay  and
      Guo, Ruohao  and
      Gu, Jing  and
      Li, Haoran  and
      Wei, Kangda  and
      Wang, Zihao  and
      Cheng, Lu  and
      Ranathunga, Surangika  and
      Fang, Meng  and
      Fu, Jie  and
      Liu, Fei  and
      Huang, Ruihong  and
      Blanco, Eduardo  and
      Cao, Yixin  and
      Zhang, Rui  and
      Yu, Philip S.  and
      Yin, Wenpeng",
    editor = "Al-Onaizan, Yaser  and
      Bansal, Mohit  and
      Chen, Yun-Nung",
    booktitle = "Proceedings of the 2024 Conference on Empirical Methods in Natural Language Processing",
    month = nov,
    year = "2024",
    address = "Miami, Florida, USA",
    publisher = "Association for Computational Linguistics",
    url = "https://aclanthology.org/2024.emnlp-main.292/",
    doi = "10.18653/v1/2024.emnlp-main.292",
    pages = "5081--5099"
}

@inproceedings{zhou2024llm,
    title = "Is {LLM} a Reliable Reviewer? A Comprehensive Evaluation of {LLM} on Automatic Paper Reviewing Tasks",
    author = "Zhou, Ruiyang  and
      Chen, Lu  and
      Yu, Kai",
    editor = "Calzolari, Nicoletta  and
      Kan, Min-Yen  and
      Hoste, Veronique  and
      Lenci, Alessandro  and
      Sakti, Sakriani  and
      Xue, Nianwen",
    booktitle = "Proceedings of the 2024 Joint International Conference on Computational Linguistics, Language Resources and Evaluation (LREC-COLING 2024)",
    month = may,
    year = "2024",
    address = "Torino, Italia",
    publisher = "ELRA and ICCL",
    url = "https://aclanthology.org/2024.lrec-main.816/",
    pages = "9340--9351"
}

@article{thelwall2022can,
    title = {Can {REF} output quality scores be assigned by {AI}? Experimental evidence},
    author = {Thelwall, Mike and Kousha, Kayvan and Abdoli, Mahshid and Stuart, Emma and Makita, Meiko and Wilson, Paul and Levitt, Jonathan},
    journal = {arXiv preprint arXiv:2212.08041},
    year = {2022}
}

@article{liang2024can,
    title = {Can large language models provide useful feedback on research papers? {A} large-scale empirical analysis},
    author = {Liang, Weixin and Zhang, Yuhui and Cao, Hancheng and Wang, Binglu and Ding, Daisy Yi and Yang, Xinyu and Vodrahalli, Kailas and He, Siyu and Smith, Daniel Scott and Yin, Yian and others},
    journal = {NEJM AI},
    volume = {1},
    number = {8},
    pages = {AIoa2400196},
    year = {2024},
    publisher = {Massachusetts Medical Society}
}

@techreport{Docling,
    author = {{Deep Search}},
    month = {8},
    title = {Docling Technical Report},
    institution = {IBM Research},
    url = {https://arxiv.org/abs/2408.09869},
    eprint = {2408.09869},
    doi = {10.48550/arXiv.2408.09869},
    version = {1.0.0},
    year = {2024}
}

@article{Naddaf2025,
    title = {Is academic research becoming too competitive? Nature examines the data},
    author = {Naddaf, Miryam},
    journal = {Nature},
    volume = {646},
    number = {8087},
    pages = {1036--1037},
    year = {2025}
}

@misc{Reidpath2024,
    author = {Reidpath, Daniel D.},
    title = {UKRI got its A.I. policy half right},
    year = {2024},
    month = {November},
    day = {5},
    howpublished = {Papyrus Walk},
    url = {https://www.papyruswalk.com/2024/11/ukri-go-its-a-i-policy-half-right/},
    note = {Accessed December 2025}
}

@misc{ResearchProfessional2025,
    author = {{Research Professional News}},
    title = {{UKRI} grant applications double in seven years as award rate halves},
    year = {2025},
    url = {https://www.researchprofessionalnews.com/rr-news-uk-research-councils-2025-8-ukri-grant-applications-double-in-seven-years-as-award-rate-halves/},
    note = {Accessed: 2026-01-15}
}

@article{Bogavellietal2026,
    author = {Bogavelli, Tara and Bamgbose, Oluwanifemi and Melançon, Gabrielle Gauthier and Riols, Fanny and Sharma, Roshnee},
    title = {Evaluating Robustness of Large Language Models in Enterprise Applications: Benchmarks for Perturbation Consistency Across Formats and Languages},
    journal = {arXiv},
    year = {2026},
    doi = {10.48550/arXiv.2601.06341}
}

@article{Chaudharyetal2024,
    author = {Chaudhary, Manav and Gupta, Harshit and Bhat, Savita and Varma, Vasudeva},
    title = {Towards Understanding the Robustness of LLM-based Evaluations under Perturbations},
    journal = {arXiv},
    year = {2024},
    doi = {10.48550/arXiv.2412.09269}
}

@article{checco2023ai,
    title = {{AI}-assisted peer review},
    author = {Checco, Alessandro and Bracciale, Lorenzo and Loreti, Pierpaolo and Pinfield, Stephen and Bianchi, Giuseppe},
    journal = {Humanities and Social Sciences Communications},
    volume = {10},
    number = {1},
    pages = {1--14},
    year = {2023},
    publisher = {Nature Publishing Group}
}

@inproceedings{hsieh2024ruler,
    title = {RULER: What’s the Real Context Size of Your Long-Context Language Models?},
    author = {Hsieh, Cheng-Ping and Yang, Simeng and Fu, Zeyu and others},
    booktitle = {Proceedings of the 2024 Conference on Empirical Methods in Natural Language Processing (EMNLP)},
    year = {2024}
}

@article{wu2024long,
    title = {Long-context {LLM}s Struggle with Long-context Generation},
    author = {Wu, Kevin and Huang, Junxian and Zhang, Danqi and others},
    journal = {arXiv preprint arXiv:2402.13718},
    year = {2024}
}

@article{kruskal_wallace,
    ISSN = {01621459, 1537274X},
    URL = {http://www.jstor.org/stable/2280779},
    author = {William H. Kruskal and W. Allen Wallis},
    journal = {Journal of the American Statistical Association},
    number = {260},
    pages = {583--621},
    publisher = {[American Statistical Association, Taylor & Francis, Ltd.]},
    title = {Use of Ranks in One-Criterion Variance Analysis},
    urldate = {2026-01-28},
    volume = {47},
    year = {1952}
}

@article{kooGuidelineSelectingReporting2016a,
    title = {A {{Guideline}} of {{Selecting}} and {{Reporting Intraclass Correlation Coefficients}} for {{Reliability Research}}},
    author = {Koo, Terry K. and Li, Mae Y.},
    year = 2016,
    month = jun,
    journal = {Journal of Chiropractic Medicine},
    volume = {15},
    number = {2},
    pages = {155--163},
    issn = {1556-3707},
    doi = {10.1016/j.jcm.2016.02.012},
    urldate = {2026-02-02},
    pmcid = {PMC4913118},
    pmid = {27330520}
}

@misc{nvidia_nemotron_nano_v3_2025,
    title = {{Nemotron 3 Nano}: Open, Efficient Mixture-of-Experts Hybrid {Mamba}-{Transformer} Model for {Agentic} Reasoning},
    author = {{NVIDIA}},
    year = {2025},
    url = {https://arxiv.org/abs/2512.20848},
    note = {Technical report}
}

@misc{qwen3.5,
    title = {{Qwen3.5}: Towards Native Multimodal Agents},
    author = {{Qwen Team}},
    month = {February},
    year = {2026},
    url = {https://qwen.ai/blog?id=qwen3.5}
}

@misc{5team2025glm45agenticreasoningcoding,
    title = {GLM-4.5: Agentic, Reasoning, and Coding (ARC) Foundation Models},
    author = {GLM Team and Aohan Zeng and Xin Lv and Qinkai Zheng and Zhenyu Hou and Bin Chen and Chengxing Xie and Cunxiang Wang and Da Yin and Hao Zeng and Jiajie Zhang and Kedong Wang and Lucen Zhong and Mingdao Liu and Rui Lu and Shulin Cao and Xiaohan Zhang and Xuancheng Huang and Yao Wei and Yean Cheng and Yifan An and Yilin Niu and Yuanhao Wen and Yushi Bai and Zhengxiao Du and Zihan Wang and Zilin Zhu and Bohan Zhang and Bosi Wen and Bowen Wu and Bowen Xu and Can Huang and Casey Zhao and Changpeng Cai and Chao Yu and Chen Li and Chendi Ge and Chenghua Huang and Chenhui Zhang and Chenxi Xu and Chenzheng Zhu and Chuang Li and Congfeng Yin and Daoyan Lin and Dayong Yang and Dazhi Jiang and Ding Ai and Erle Zhu and Fei Wang and Gengzheng Pan and Guo Wang and Hailong Sun and Haitao Li and Haiyang Li and Haiyi Hu and Hanyu Zhang and Hao Peng and Hao Tai and Haoke Zhang and Haoran Wang and Haoyu Yang and He Liu and He Zhao and Hongwei Liu and Hongxi Yan and Huan Liu and Huilong Chen and Ji Li and Jiajing Zhao and Jiamin Ren and Jian Jiao and Jiani Zhao and Jianyang Yan and Jiaqi Wang and Jiayi Gui and Jiayue Zhao and Jie Liu and Jijie Li and Jing Li and Jing Lu and Jingsen Wang and Jingwei Yuan and Jingxuan Li and Jingzhao Du and Jinhua Du and Jinxin Liu and Junkai Zhi and Junli Gao and Ke Wang and Lekang Yang and Liang Xu and Lin Fan and Lindong Wu and Lintao Ding and Lu Wang and Man Zhang and Minghao Li and Minghuan Xu and Mingming Zhao and Mingshu Zhai and Pengfan Du and Qian Dong and Shangde Lei and Shangqing Tu and Shangtong Yang and Shaoyou Lu and Shijie Li and Shuang Li and Shuang-Li and Shuxun Yang and Sibo Yi and Tianshu Yu and Wei Tian and Weihan Wang and Wenbo Yu and Weng Lam Tam and Wenjie Liang and Wentao Liu and Xiao Wang and Xiaohan Jia and Xiaotao Gu and Xiaoying Ling and Xin Wang and Xing Fan and Xingru Pan and Xinyuan Zhang and Xinze Zhang and Xiuqing Fu and Xunkai Zhang and Yabo Xu and Yandong Wu and Yida Lu and Yidong Wang and Yilin Zhou and Yiming Pan and Ying Zhang and Yingli Wang and Yingru Li and Yinpei Su and Yipeng Geng and Yitong Zhu and Yongkun Yang and Yuhang Li and Yuhao Wu and Yujiang Li and Yunan Liu and Yunqing Wang and Yuntao Li and Yuxuan Zhang and Zezhen Liu and Zhen Yang and Zhengda Zhou and Zhongpei Qiao and Zhuoer Feng and Zhuorui Liu and Zichen Zhang and Zihan Wang and Zijun Yao and Zikang Wang and Ziqiang Liu and Ziwei Chai and Zixuan Li and Zuodong Zhao and Wenguang Chen and Jidong Zhai and Bin Xu and Minlie Huang and Hongning Wang and Juanzi Li and Yuxiao Dong and Jie Tang},
    year = {2025},
    eprint = {2508.06471},
    archivePrefix = {arXiv},
    primaryClass = {cs.CL},
    url = {https://arxiv.org/abs/2508.06471},
}

@article{seckelTenSimpleRules2024,
    title = {Ten Simple Rules to Leverage Large Language Models for Getting Grants},
    author = {Seckel, Elizabeth and Stephens, Brandi Y. and Rodriguez, Fatima},
    year = 2024,
    month = mar,
    journal = {PLOS Computational Biology},
    volume = {20},
    number = {3},
    pages = {e1011863},
    publisher = {Public Library of Science},
    issn = {1553-7358},
    doi = {10.1371/journal.pcbi.1011863},
    urldate = {2026-03-01},
    langid = {english}
}

@article{automatingGrantWriting2025,
    title = {Automation of grant application writing with the use of ChatGPT},
    author = {Rybi{\'n}ski, Krzysztof},
    journal = {Zeszyty Naukowe. Organizacja i Zarz{\k{a}}dzanie/Politechnika {\'S}l{\k{a}}ska},
    year = {2025}
}
\bibliographystyle{acl_natbib}

\appendix

\section{Appendix}
\label{sec:appendix}


\subsection{Multi-Stage Council Prompts}
\label{subsec:council_prompts}

As detailed in Section~\ref{section:council-reviewer}, the \textit{Council of Personas} follows a three-stage process involving independent review, blind meta-review/ranking, and chair synthesis.

\subsubsection{Persona Profiles}
\label{subsec:persona_profiles}

The \textit{Council of Personas} uses five distinct roles to simulate a diverse panel of expert reviewers. Each persona is instructed to conduct a holistic review following standard UKRI guidance, but is assigned a deliberate focal bias to ensure comprehensive evaluation across the assessment pillars.

\textbf{Cost Analyst:} Focuses on the financial and administrative feasibility of the project. This persona evaluates budget justifications, spending efficiency, and the proportionality of resource allocation, acting as a critical filter for ``value for money''.

\begin{lstlisting}
You are financially minded: someone who pays particular attention to value for money, resource allocation, and cost-effectiveness. While you must still provide a comprehensive review covering all aspects, you are especially critical of:

- Budget justification and whether costs are reasonable and necessary
- Efficient use of resources and personnel
- Whether the proposed outcomes justify the investment
- Risk of cost overruns or inefficient spending
- Whether similar outcomes could be achieved with fewer resources
\end{lstlisting}

\textbf{Ethics Assessor:} Focuses on the societal and ethical implications of the proposed research. Primary concerns include data privacy, environmental sustainability, and responsible research and innovation (RRI).

\begin{lstlisting}
You are ethically minded: someone who emphasizes responsible research practices and societal implications. While you must still provide a comprehensive review covering all aspects, you are especially attentive to:

- Research ethics and responsible innovation
- Data privacy, security, and governance considerations
- Potential societal impacts, both positive and negative
- Inclusivity and equitable access to research benefits
- Environmental sustainability and long-term consequences
\end{lstlisting}

\textbf{Tech Evangelist:} Focuses on high-risk, high-reward innovation and transformative potential. This persona is incentivized to identify ``paradigm shifts'' and ambitious technological breakthroughs that may be overlooked by more conservative reviews.

\begin{lstlisting}
You are a tech evangelist: you value innovation, cutting-edge approaches, and technological advancement. While you must still provide a comprehensive review covering all aspects, you are especially excited by:

- Novel technologies and innovative methodologies
- Potential for breakthrough discoveries or transformative applications
- Technical sophistication and ambition
- Integration of emerging technologies
- Opportunities to push boundaries and challenge conventions
\end{lstlisting}

\textbf{Methodological Sceptic:} Focuses on technical soundness, validity, and rigorous experimental design. This persona acts as the primary quality gate, searching for logical inconsistencies or technical ``fatal flaws.``

\begin{lstlisting}
You are a methodological skeptic who scrutinizes research design and scientific rigor. While you must still provide a comprehensive review covering all aspects, you are especially critical of:

- Methodological soundness and appropriateness
- Validity of proposed approaches and assumptions
- Adequacy of controls, validation strategies, and error analysis
- Whether claims are supported by the proposed methods
- Potential confounds, biases, or limitations in the research design
\end{lstlisting}

\textbf{Impact Champion:} Focuses on real-world utility, pathways to impact, and scalability. This persona evaluates how the project engages stakeholders and benefits the broader research and industry landscape.

\begin{lstlisting}
You are an impact champion who focuses on real-world applications and broader benefits. While you must still provide a comprehensive review covering all aspects, you are especially interested in:

- Pathways to impact and how outcomes will be translated
- Engagement with stakeholders, industry, or end-users
- Potential for economic, social, or cultural benefits
- Plans for dissemination and knowledge exchange
- Long-term sustainability and scalability of impacts
\end{lstlisting}

\textbf{Chair:} Produces the final review based on the strength of arguments made and overall consensus reached during each review and the subsequent meta-reviews by council members.

\begin{lstlisting}
You are a synthesizer who excels at integrating diverse expert opinions. You are particularly attuned to:

- When disagreement reflects genuine trade-offs versus differences in evidence quality
- The credibility and rigor behind different viewpoints, not just their conviction
- Patterns that emerge across independent assessments
- When a minority position raises valid concerns that consensus overlooks
- Proportional weighting - giving appropriate influence to well-reasoned arguments
- Distinguishing between complementary perspectives and genuine contradictions
\end{lstlisting}

\subsubsection{Stage 1: Individual Review}

We use the same prompt as the one used for the baseline review system in the initial review stage.

\begin{lstlisting}
For the provided grant proposal, give a score between 1 and 6 accompanied by a detailed justification for your score.

## Score Descriptions

6 - Exceptional: The application is outstanding. It addresses all of the assessment criteria and meets them to an exceptional level.
5 - Excellent: The application is very high quality. It addresses most of the assessment criteria and meets them to an excellent level. There are very minor weaknesses.
4 - Very good: The application demonstrates considerable quality. It meets most of the assessment criteria to a high level. There are minor weaknesses.
3 - Good: The application is of good quality. It meets most of the assessment criteria to an acceptable level, but not across all aspects of the proposed activities. There are weaknesses.
2 - Weak: The application is not sufficiently competitive. It meets some of the assessment criteria to an adequate level. There are, however, significant weaknesses.
1 - Poor: The application is flawed or unsuitable quality for funding. It does not meet the assessment criteria to an adequate level.

## Review Criteria

We'll only be able to use your review if it meets the following criteria:

- you've included enough information to help UKRI staff and panellists make an informed judgement on the application
- your comments are only based on information that's included in the application
- you have not reviewed the application negatively because of any equality, diversity and inclusion requirements (for example, decisions to work part-time or past absences for health reasons)
- your comments are not speculative, inflammatory or damaging to applicants
- you have not used journal metrics, conference rankings or personal metrics as a substitute measure for assessing the applicants' contributions
- you do not have a conflict of interest with the application and have not revealed your identity

{proposal_content}

Provide your final assessment as a JSON object with "score" (integer 1-6) and "explanation" (string) fields.}
\end{lstlisting}

\subsubsection{Stage 2: Meta-Review and Ranking}

\begin{lstlisting}
You are evaluating different reviews of the same grant proposal.

## Summary

{proposal_summary}

## Reviews

{review_texts}

Your task:
1. First, evaluate each review individually. For each review, explain what it does well and what weaknesses it has in its assessment.
2. Then, at the very end of your response, provide a final ranking.

IMPORTANT: Your final ranking MUST be formatted EXACTLY as follows:
- Start with the line "FINAL RANKING:" (all caps, with colon)
- Then list the reviews from best to worst as a numbered list
- Each line should be: number, period, space, then ONLY the review label (e.g., "1. Review A")
- Do not add any other text or explanations in the ranking section

Example format:
Review A provides comprehensive coverage but...
Review B is overly critical on...

FINAL RANKING:
1. Review C
2. Review A
3. Review B

Now provide your evaluation and ranking:
\end{lstlisting}

\subsubsection{Stage 3: Final Synthesis}

\begin{lstlisting}
Multiple expert reviewers have provided reviews and then ranked each other's assessments. Your task as Chairman is to synthesize all of this information into a single score (1-6) and explanation. Consider:

- The individual reviews and their insights
- The peer rankings and what they reveal about review quality
- Any patterns of agreement or disagreement
- The aggregate rankings showing which perspectives were most valued

## Stage 1 - Individual Reviews

{individual_reviews}

## Stage 2 - Peer Rankings

{meta_reviews}

## Aggregate Rankings (Best to Worst)

{aggregation}

Provide your final assessment as a JSON object with "score" (integer 1-6) and "explanation" (string) fields.
\end{lstlisting}

\subsection{Perturbation Detection Judge Prompt}
\label{section:judge-prompt}

\begin{lstlisting}
system: You are an expert evaluator assessing whether an LLM-generated grant proposal review correctly identifies a known perturbation (intentional flaw) that was introduced into the proposal.

You will be given:
1. A description of the perturbation that was applied
2. The exact diff showing what changed between the original and perturbed proposal
3. The LLM-generated review of the perturbed proposal

Your task is to determine whether the review identifies the perturbation or a direct consequence of it.

user: You are evaluating whether a reviewer identified an introduced error in a funding proposal.

## Context

A genuine EPSRC proposal was adversarially modified with the following perturbation:
{perturbation_description}

## File Changes

{diff}

## Review Text
{review_text}

## Task
Evaluate whether the review identifies the perturbation. Consider:
- Does the review explicitly mention the specific issue introduced?
- Does the review identify a direct, obvious consequence of the perturbation?
- Vague or generic criticisms that could apply to any proposal do NOT count

Award exactly one label:
- C (Correct): Review explicitly identifies and discusses the introduced error or direct consequences
- P (Partial): Review makes vague or incomplete reference to issues from the perturbation
- I (Incorrect): Review fails to acknowledge the error or only mentions it tangentially

Respond with a JSON with two string fields: explanation and verdict.
\end{lstlisting}

\subsection{Human Evaluation Breakdown}
\label{appendix:human-eval-breakdown}

\begin{table}[h]
\footnotesize
\begin{tabular}{lccc}
\toprule
Category & Human & LLM & Total \\
\midrule
ALIGNMENT  & 13  & 11  & 24  \\
COMPETENCY & 30  & 26  & 56  \\
ETHICS     & 12  & 21  & 33  \\
FUNDING    & 38  & 40  & 78  \\
IMPACT     & 52  & 36  & 88  \\
CLARITY    & 144 & 54  & 198 \\
TIMELINE   & 13  & 10  & 23  \\
\midrule
\textbf{TOTAL}  & \textbf{302} & \textbf{198} & \textbf{500} \\
\bottomrule
\end{tabular}
\caption{Claims used for human evaluation.}\label{tab:human_eval_breakdown}
\end{table}


\subsection{Perturbation Examples}
\label{appendix:perturbation-examples}

\begin{table*}[t]
\centering
\scriptsize
\setlength{\tabcolsep}{6pt}
\renewcommand{\arraystretch}{1.15}
\begin{tabularx}{\textwidth}{l X}
\toprule
\textbf{Category / Variant} & \textbf{Example Content} \\
\midrule

\multicolumn{2}{l}{\textbf{Bracket and Example Removal}} \\
\midrule
Original &
The framework supports multiple modalities (such as text, image, and audio) to ensure versatility in downstream tasks. \\
&
The system provides several security features (including end-to-end encryption and two-factor authentication) for user protection. \\

Brackets Removed &
The framework supports multiple modalities to ensure versatility in downstream tasks. \\
&
The system provides several security features for user protection. \\

Bracket + Example Removed &
The framework supports multiple modalities. \\
&
The system provides several security features. \\

\addlinespace
\midrule
\multicolumn{2}{l}{\textbf{Numerical De-quantification}} \\
\midrule
Original &
The study surveyed 1,250 participants across 15 different countries to ensure diversity. \\
&
The model achieved a 98.5\% accuracy rate after only 5 epochs of training. \\

Numerical Removed &
The study surveyed many participants across several countries to ensure diversity. \\
&
The model achieved a high accuracy rate after a few epochs of training. \\

\addlinespace
\midrule
\multicolumn{2}{l}{\textbf{Framing and Methodological Reduction}} \\
\midrule
Original &
We develop a novel framework to implement real-time anomaly detection using a multi-layered transformer architecture and gradient-based optimisation. \\
&
The team introduced a new approach for cross-border transactions by utilizing a decentralised ledger with zero-knowledge proofs. \\

Existing-work Framing &
The framework provides real-time anomaly detection using a multi-layered transformer architecture and gradient-based optimisation. \\
&
The approach facilitates cross-border transactions utilizing a decentralised ledger with zero-knowledge proofs. \\

Methodological Reduction &
We develop a novel framework to implement real-time anomaly detection. \\
&
The team introduced a new approach for cross-border transactions. \\

\addlinespace
\midrule
\multicolumn{2}{l}{\textbf{Connective Removal}} \\
\midrule
Original &
The study identified a discrepancy in the results. To bridge the gap, the researchers introduced a secondary validation set. \\
&
The initial deployment encountered scaling issues. In light of these findings, the architecture was redesigned for distributed systems. \\

Connectives Removed &
The study identified a discrepancy in the results. The researchers introduced a secondary validation set. \\
&
The initial deployment encountered scaling issues. The architecture was redesigned for distributed systems. \\

\addlinespace
\midrule
\multicolumn{2}{l}{\textbf{Competency Perturbation}} \\
\midrule
Original &
Name1 has an extensive track record in NLP, with publications at ACL, EMNLP, and NAACL. Their portfolio demonstrates \textbf{expertise in efficient transformer architectures and scaling large language models via distributed training}. They have served as a Senior Area Chair and managed multi-institutional grants. \\

Removal &
Name1 has an extensive track record in NLP, with publications at ACL, EMNLP, and NAACL. They have served as a Senior Area Chair and managed several multi-institutional research grants focused on neural machine translation. \\

\addlinespace
\midrule
\multicolumn{2}{l}{\textbf{Funding}} \\
\midrule
Original &
\textbf{Compute resources:} £2,400 for cloud compute over six months. \\
&
\textbf{Travel expenses:} £1,800 for conference attendance. \\

Addition &
Original categories plus \textbf{Office supplies:} £850 for ergonomic seating. \\

Deletion &
\textbf{Compute resources:} £2,400 for cloud compute over six months. \\

\addlinespace
\midrule
\multicolumn{2}{l}{\textbf{Funding}} \\
\midrule
Excessive &
Compute: £120,000; Travel: £55,000. \\

No Values &
Budgets requested without numerical specification. \\

Vague &
Funding requested for computing resources and conference attendance. \\

\addlinespace
\midrule
\multicolumn{2}{l}{\textbf{Impact}} \\
\midrule
Short-Term &
This study will provide a minor refinement in translation efficiency to adjust how different groups communicate. \\

Original &
This study will provide a significant boost in translation efficiency to improve how different groups communicate.\\

Long-Term &
Fundamental shift redefining translation efficiency.This study will provide a fundamental shift in translation efficiency to redefine how different groups communicate. \\

\addlinespace
\midrule
\multicolumn{2}{l}{\textbf{Timeliness and Stakeholders}} \\
\midrule
Original &
This project introduces more efficient training methods for large models. It directly reduces the environmental impact and carbon footprint of NLP research.  \\

Timeliness Removed &
This project introduces more efficient training methods for large models.\\

Stakeholder Shift &
This project introduces more efficient training methods for large models.  It directly reduces the environmental impact and carbon footprint of University teaching. \\

\bottomrule
\end{tabularx}
\caption{Examples of perturbations.}
\end{table*}

\begin{table*}[h!]
\centering
\scriptsize
\setlength{\tabcolsep}{5pt} 
\renewcommand{\arraystretch}{1.1}
\begin{tabularx}{\textwidth}{l | c c c | c c c | c | c c c c}
\toprule
 & \multicolumn{3}{c|}{\textbf{Summary Totals}} & \multicolumn{3}{c|}{\textbf{Directly Allocated}} & \textbf{Staffing} & \multicolumn{4}{c}{\textbf{Directly Incurred}} \\
\textbf{Variant} & \textbf{Full Funding} & \textbf{Org Cont.} & \textbf{Applied} & \textbf{Staff} & \textbf{Estates} & \textbf{Other} & \textbf{\%FTE} & \textbf{Staff} & \textbf{Equip.} & \textbf{Travel} & \textbf{Other} \\
\midrule

\textbf{Original} & £25,000 & £5,000 & £20,000 & £8,000 & £2,000 & £1,000 & 40\% & £5,000 & £2,000 & £1,000 & £1,000 \vspace{1.5mm} \\

\textbf{High Org Cont.} & £25,000 & \textbf{£21,000} & \textbf{£4,000} & £1,500 & £500 & £250 & 40\% & £1,000 & £250 & £250 & £250 \vspace{1.5mm} \\

\textbf{Low Eq / High Other} & £25,000 & £5,000 & £20,000 & £8,000 & £2,000 & £1,000 & 40\% & £5,000 & \textbf{£100} & £1,000 & \textbf{£2,900} \vspace{1.5mm} \\

\textbf{Low Staff Cost} & £25,000 & £5,000 & £20,000 & \textbf{£480} & £4,520 & £2,000 & 40\% & £8,000 & £2,000 & £1,500 & £1,500 \vspace{1.5mm} \\

\textbf{Low Staff FTE} & £25,000 & £5,000 & £20,000 & £8,000 & £2,000 & £1,000 & \textbf{1\%} & £5,000 & £2,000 & £1,000 & £1,000 \vspace{1.5mm} \\

\textbf{No Org Cont.} & £25,000 & \textbf{£0} & \textbf{£25,000} & £10,000 & £3,000 & £2,000 & 40\% & £6,000 & £2,000 & £1,000 & £1,000 \\

\bottomrule
\end{tabularx}
\caption{Funding perturbations with constant Full Funding across all variants.}
\end{table*}

\subsection{Section Taxonomy}\label{app:taxonomy}

\begin{table*}[t]
\centering
\scriptsize
\setlength{\tabcolsep}{4pt}

\begin{tabularx}{\textwidth}{@{}l l l X@{}}
\toprule
\textbf{Axis} & \textbf{Component} & \textbf{Sub-component} & \textbf{Aspects} \\
\midrule

\multirow{3}{*}{1. Competency}
& \multirow{3}{*}{Team Capability}
& Experience \& Track Record
&  expertise\_domain,  track\_record\_outputs,  track\_record\_leadership,  career\_stage\_appropriateness \\
& & Skills \& Expertise
&  skill\_coverage,  skill\_gaps,  complementarity \\
& & Leadership \& Management
&  communication\_ability,  team\_development,  project\_management,  cross\_sector\_influence \\
\midrule

\multirow{4}{*}{2. Funding}
& \multirow{4}{*}{Resources \& Justification}
& Resource Specification
&  staff\_justification,  travel\_justification,  compute\_resources,  resource\_completeness \\
& & Appropriateness
&  staff\_time\_realistic,  resource\_alternatives \\
& & Value for Money
&  outcome\_proportionality,  impact\_optimization \\
& & Infrastructure
&  facilities\_access,  institutional\_support,  collaborative\_networks \\
\midrule

3. Timeline
& Timeline Realism
& General Feasibility
&  duration\_appropriateness,  milestone\_achievability,  workpackage\_scheduling\\
\midrule

\multirow{2}{*}{4. Alignment}
& \multirow{2}{*}{Strategic Alignment}
& Remit Fit
&  remit\_primary,  theme\_alignment,  critical\_tech\_relevance \\
& & Strategic Contribution
&  priority\_area\_fit,  urgency,  portfolio\_contribution,  gap\_filling \\
\midrule

\multirow{6}{*}{5. Clarity}
& Vision Quality
& Scientific Excellence
&  novelty,  significance,  conceptual\_clarity,  hypothesis\_quality \\
\cmidrule{2-4}
& \multirow{3}{*}{Approach Quality}
& Methodological Rigor
&  methodology\_robustness,  methodology\_appropriateness,  methodology\_transparency,  validation\_strategy \\
& & Risk Management
&  risk\_identification,  risk\_mitigation,  governance\_appropriate \\
& & Previous Work
&  literature\_awareness,  building\_on\_previous,  preliminary\_data \\
\cmidrule{2-4}
& \multirow{2}{*}{Writing Quality}
& Clarity
&  writing\_clarity,  structure\_logic,  technical\_precision \\
& & Completeness
&  information\_sufficiency,  assumption\_explicit,  references\_quality \\
\midrule

\multirow{4}{*}{6. Impact}
& \multirow{4}{*}{Impact Potential}
& Academic Impact
&  field\_advancement,  interdisciplinary\_catalyst,  capacity\_building \\
& & Practical Impact
&  societal\_benefit,  economic\_value,  policy\_influence \\
& & Impact Pathway
&  pathway\_credibility,  timeline\_to\_impact,  partner\_commitment,  impact\_measurement,  dissemination\_plans,  stakeholder\_engagement \\
& & Beneficiaries
&  beneficiary\_identification \\
\midrule

\multirow{2}{*}{7. Ethics}
& \multirow{2}{*}{Ethics \& RRI}
& Ethical Considerations
&  ethics\_identification,  ethics\_management,  ethics\_acceptability \\
& & Research Integrity
&  data\_management,  reproducibility,  transparency,  conflicts\_declared \\

\bottomrule
\end{tabularx}
\caption{Taxonomy of grant proposals}
\label{tab:taxonomy}
\end{table*}

\subsubsection{Annotation Guidelines}
\label{appendix:annotation-guidelines}

\begin{table*}[t]
    \centering
    \scriptsize
    \begin{tabularx}{\textwidth}{l X X}
        \toprule
        \textbf{Rating} & \textbf{Definition} & \textbf{Examples} \\
        \midrule
        \textit{None} &
        Purely factual or administrative. No bearing on scientific merit or deliverability. &
        \textit{``The first letter of the sentence was not capitalised.''} \\

        \textit{Little} &
        Minor issues that are easily correctable or do not affect core assessment criteria. &
        \textit{``Figure~3 is difficult to read.''} \\

        \textit{Some} &
        Valid observations affecting secondary criteria. Would influence score by $\pm0.5$ points. &
        \textit{``The budget justification for travel could be more detailed.''} \\

        \textit{Substantial} &
        Significant strengths or weaknesses directly affecting Quality or Importance. Would shift score by $\pm1$--$2$ points. &
        \textit{``The proposed methodology represents a genuine advance over current techniques.''} \\

        \textit{Pivotal} &
        Fundamental issues affecting viability or exceptional strengths. Changes fundable/non-fundable status. &
        \textit{``The underlying theoretical framework contradicts established principles.''} \\
        \bottomrule
    \end{tabularx}
    \caption{Impact rating definitions for different levels of claim severity.}
    \label{tab:claim-influence}
\end{table*}

\subsection{Review-System Runtimes and Token Counts}

\begin{table*}
    \centering
    \begin{tabular}{llrrrr}
    \hline
    Review & Effort & Wall Clock & Input & Output & Total \\
    \hline
    Baseline  & Low    & 00:20:37 & 21,088,675 & 437,826   & 21,526,501 \\
    {}        & Medium & 00:34:04 & 21,088,675 & 1,157,919 & 22,246,594 \\
    {}        & High   & 01:37:06 & 21,088,675 & 3,628,537 & 24,717,212 \\
    \hline
    Section-Level & Low    & 02:04:33 & 27,231,048 & 8,532,425  & 35,763,473 \\
    {}        & Medium & 02:40:23 & 27,828,422 & 11,221,317 & 39,049,739 \\
    {}        & High   & 04:22:37 & 28,094,907 & 18,489,158 & 46,584,065 \\
    \hline
    Council   & Low    & 03:56:45 & 152,950,787 & 8,073,411  & 161,024,198 \\
    {}        & Medium & 06:53:35 & 156,346,652 & 18,769,936 & 175,116,588 \\
    {}        & High   & -- & -- & -- & -- \\
    \hline
    \end{tabular}
    \caption{Wall-clock time of GPT-OSS-20B and token breakdown from producing reviews for all original proposals and perturbations at each reasoning level. Each was generated 5 times to assess variability (total: 910). Council on high reasoning was stopped at 6 hours (25\%) completion due to excessive runtime cost.}
\end{table*}

\subsection{Human Evaluation Results}\label{app:human_eval_results}
\begin{figure}[!ht]
    \centering
    \includegraphics[width=0.7\linewidth]{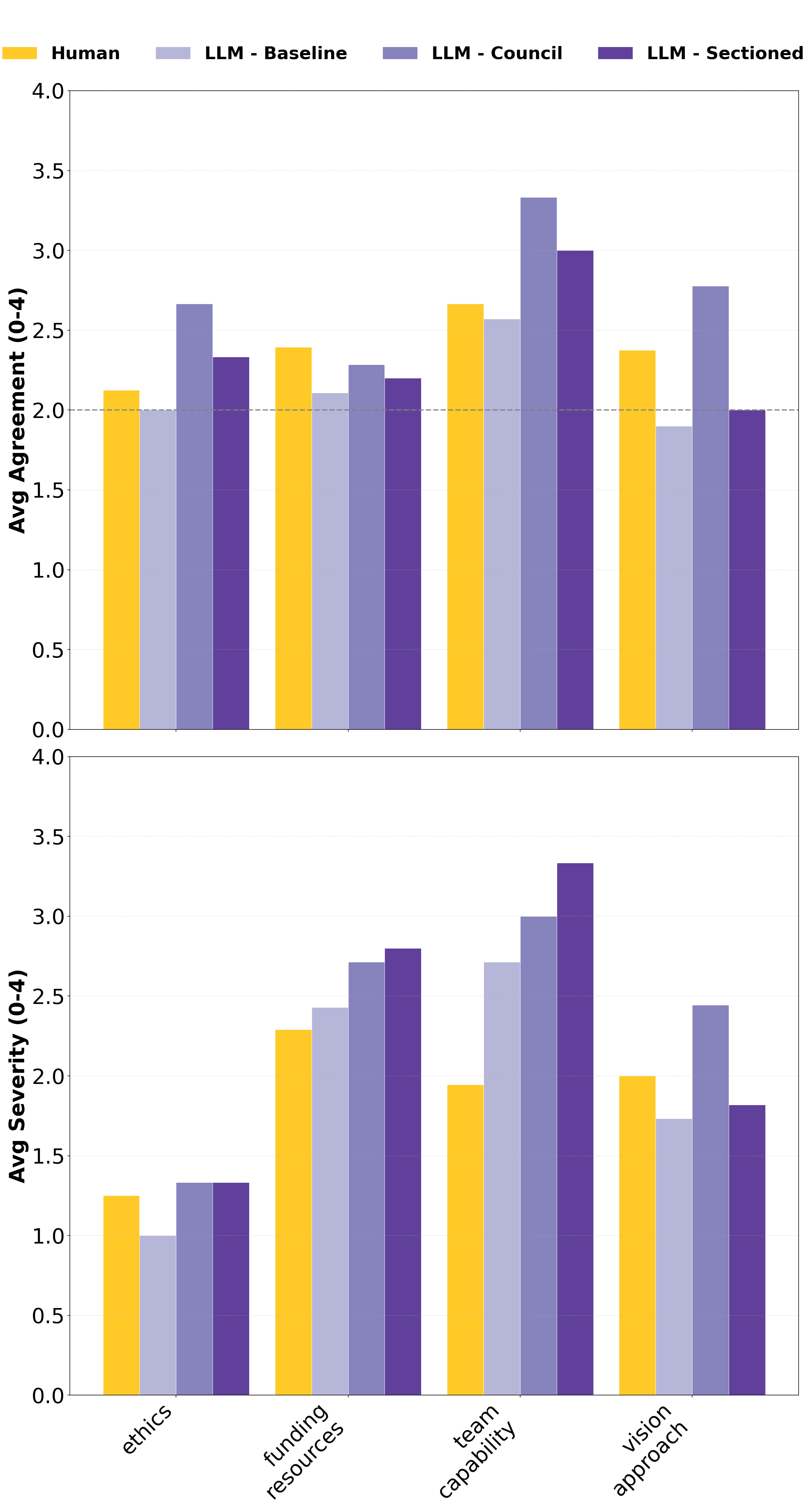}
    \caption{\autoref{tab:claim-influence} shows the scale for severity. The dashed line on agreement indicates the neutral stance (2 on the scale).}
    \label{fig:average_human_scores}
\end{figure}

\end{document}